%% file: main.tex
\documentclass{article}
\input{preamble}
\title{Beyond the Point Mass: Acceptance-Derived\\Weights and Distributional Losses\\for Speculative Decoding}
\author{%
\normalsize Tianyu Wu\textsuperscript{1,*}\quad Zhuohan Wang\textsuperscript{1,*}\quad
Jonathan Geuter\textsuperscript{1,*}\quad Yu Yao\textsuperscript{2,*}\\[0.4em]
\normalsize Weifan Jiang\textsuperscript{1}\quad Chengxi Zhang\textsuperscript{1}\quad
Ju Li\textsuperscript{2}\quad Minlan Yu\textsuperscript{1}\\[0.6em]
\small \textsuperscript{1}Harvard University\qquad
\textsuperscript{2}Massachusetts Institute of Technology\\[0.4em]
\small \textsuperscript{*}Equal contribution.
}
\begin{document}
\raggedbottom
\brokenpenalty=10000
\maketitle
\begin{center}
\small Code: \url{https://github.com/Lucas-TY/D-PAL-SpecForge}
\end{center}
\begin{abstract}
In speculative decoding, a lightweight draft model proposes tokens that the target model verifies in parallel, speeding up generation without changing the target distribution. \dpace{}~\citep{wu2026dpace} trains such drafters with a loss that weights each position by its contribution to the expected accepted length. Both its weights and its cross-entropy (CE) loss use only the draft's probability of the recorded target token, treating the target as a point mass. Under rejection sampling, however, a sampled draft token is accepted with probability equal to the overlap of the draft and target distributions, one minus their total variation (TV). We derive \textit{\dpalname{} (Dynamic Position-wise Acceptance Loss)} from the exact expected accepted length, using this overlap as the acceptance signal. This yields overlap-based position weights with log acceptance (LA) as the per-position loss, and the objective reduces to \dpace{} for a point-mass target. Since the weights are detached, any per-position loss can replace LA in practice. We show that forward KL divergence and R\'enyi divergence of order $1/2$ match or exceed LA, and that R\'enyi-$1/2$ provably bounds the true acceptance rate from above and below. Replacing CE with R\'enyi-$1/2$ improves acceptance length under both \dpace{}'s weights and ours. On six benchmarks with a Qwen3-4B target, \dpalname{} raises mean acceptance length over \dpace{} from 4.50 to 4.57 at $T=0$ and from 4.20 to 4.28 at $T=1$, and consistently achieves higher throughput at both temperatures.

\end{abstract}

\section{Introduction}
Speculative decoding accelerates autoregressive generation without changing the target model's output distribution~\citep{leviathan2023fast,chen2023speculative}: a lightweight draft model proposes tokens, and the target model verifies them in one parallel pass. DFlash drafts a whole block in one forward pass~\citep{chen2026dflash}. Since verification stops at the first rejection, early block positions matter most. DFlash trains with cross-entropy (CE) and a fixed position decay, and DSpark and H-Spec add a total variation (TV) term to CE under the same decay~\citep{cheng2026dspark,jiang2026hspec}. LK Losses optimizes acceptance directly, with the negative log acceptance or an adaptive mix of Kullback--Leibler (KL) divergence and TV, but still uses fixed decay~\citep{samarin2026lk}. \dpace{} (Dynamic Position-Aware Cross-Entropy) instead derives position weights from a surrogate of the expected accepted length~\citep{wu2026dpace}. SpecForge~\citep{specforge2026dpaceconfig}, Speculators~\citep{speculatorstrain2026}, and TorchSpec~\citep{torchspec2026dflash} use it to train DFlash drafters.

\dpace{} treats the target as a point mass on the recorded token $y^\star_t$: it estimates acceptance by the draft's confidence $q_t(y^\star_t)$ and fits that token with CE. Under standard rejection sampling, however, acceptance depends on the full target distribution $p_t$ through the overlap $\widehat\alpha_t=\sum_v\min(p_{t,v},q_{t,v})=1-\TV(p_t,q_t)$~\citep{leviathan2023fast}. When the target is uncertain, confidence can underestimate this overlap by a wide margin. For example, if $p_t=q_t$ is uniform over $k$ tokens, the verifier always accepts a sampled draft token ($\widehat\alpha_t=1$), yet $q_t(y^\star_t)=1/k$. CE likewise fits only the recorded token.

We propose \textbf{\dpalname{}} (\textbf{D}ynamic \textbf{P}osition-wise \textbf{A}cceptance \textbf{L}oss), which we derive by differentiating the expected accepted length with the \textit{true} acceptance signal $1-\TV(p_t,q_t)$. The derivation yields overlap-based position weights paired with a log acceptance (LA) loss, $-\log\widehat\alpha_t$, as the local loss, which reduces to \dpace{} when $p_t$ is a point mass. Because \dpalname{} detaches the weights, we can swap LA for any local loss in practice. We write \dpal{X} for \dpalname{} with local loss X. We show overlap bounds the acceptance of the argmax proposals that deployed DFlash uses (Proposition~\ref{prop:argmax}). Replacing CE with a distributional loss, R\'enyi divergence of order $1/2$, raises mean acceptance length under both confidence and overlap weights. In our experiments, we show that LA is not the best local loss. It is not differentiable where a draft probability equals the target probability, and its gradient with respect to the draft logits does not shrink near $q_t=p_t$, whereas the KL and R\'enyi gradients vanish there. Forward KL and R\'enyi-$1/2$ match or exceed LA, and R\'enyi-$1/2$, unlike KL, bounds overlap from both sides (Lemma~\ref{thm:overlap}). We also test TV-based losses, where pure TV collapses and CE+TV falls short, and analyze their gradients. 

\Needspace{6\baselineskip}
Our contributions are:

\begingroup
\interlinepenalty=10000
\renewcommand{\labelenumi}{\roman{enumi})}
\begin{enumerate}
\setlength{\itemsep}{4pt}
\setlength{\parskip}{0pt}
\item \textbf{Derivation from expected accepted length.} Differentiating the exact expected accepted length with the overlap as the acceptance signal gives overlap-based weights paired with LA. Proposition~\ref{prop:argmax} connects the overlap to the argmax verification used in deployment. We show \dpalLA{} improves on \dpace{}, raising mean acceptance length on Qwen3-4B from 4.50 to 4.54 at $T=0$ and from 4.20 to 4.24 at $T=1$ (Table~\ref{tab:results}).

\item \textbf{Local loss beyond log acceptance.} The derivation points to LA, but LA is not differentiable where the draft matches the target. Keeping the same weights, we replace it with forward KL and with R\'enyi-$1/2$, and both reach the same or higher acceptance length (Table~\ref{tab:results}). We also show R\'enyi-$1/2$ bounds overlap from above and below (Lemma~\ref{thm:overlap}).

\item \textbf{Consistent gains over \dpace{} and DFlash.} In experiments across temperatures, block sizes, draft depths, target models at the 4B and 8B scale, and thinking modes, we show that \textbf{\dpalname{} consistently achieves longer acceptance lengths and higher throughputs than both \dpace{} (up to 5\%) and DFlash (up to 11\%).}


\end{enumerate}
\endgroup

\section{Background}

\subsection{Speculative decoding with DFlash}
\label{sec:bg-dflash}
In speculative decoding~\citep{leviathan2023fast,chen2023speculative}, the target model scores a block of $B$ draft proposals in one forward pass and accepts a contiguous prefix of length $\tau$.

DFlash~\citep{chen2026dflash} drafts with a small block-diffusion model that projects hidden states from a few target layers into the keys and values of every draft layer. A block of $B$ positions starts with an anchor, the last verified token, followed by $D=B-1$ mask tokens. In one forward pass, the drafter predicts distributions $q_1,\dots,q_D$ for all masked positions. Each $q_t$ thus depends on the prefix but not on earlier proposals in the block, while the target's $p_t$ conditions on all tokens before $t$.

In training, DFlash samples anchor positions in responses generated by the frozen target and predicts the next $D$ recorded tokens $y^\star_1,\dots,y^\star_D$ from each anchor context. Since a proposal counts only if every earlier one is accepted, DFlash fits $y^\star_t$ with CE and weights positions by a fixed exponential decay,
\begin{equation}
\mathcal{L}_{\text{DFlash}}=\sum_{t=1}^{D} e^{-(t-1)/\gamma}\,\bigl[-\log q_t(y^\star_t)\bigr],
\label{eq:static}
\end{equation}
with $\gamma=7$ for $B=16$. The frozen target output head also gives the full target distribution $p_t$ at each position, but DFlash does not use it.

\subsection{Expected accepted length}\label{sec:position-training}
Let $L=\tau-1$ count the accepted draft tokens, and let $\alpha_t=\Pr(L\ge t\mid L\ge t-1)$ be the conditional acceptance probability at position $t$. The tail-sum identity gives the exact expected accepted length~\citep{li2026bebop,wu2026dpace}
\begin{equation}
 \E[L]=\sum_{k=1}^{D}\prod_{i=1}^{k}\alpha_i.
 \label{eq:expected-length}
\end{equation}
Each $\alpha_t$ appears in every accepted prefix ending at $k\ge t$. For positive $\alpha_t$, its relative contribution and the gradient of $\E[L]$ are
\begin{equation}
 w_t^{\rm raw}=\alpha_t\frac{\partial\E[L]}{\partial\alpha_t}
 =\sum_{k=t}^{D}\prod_{i=1}^{k}\alpha_i,
 \qquad \nabla\E[L]=\sum_{t=1}^{D}w_t^{\rm raw}\nabla\log\alpha_t.
 \label{eq:relative-credit}
\end{equation}
Every term is positive, so $w^{\rm raw}_t$ decreases with $t$.

\subsection{Position weights and local losses}\label{sec:dpace-background}
Equation~\eqref{eq:relative-credit} suggests drafter objectives for a block of the form $\sum_{t=1}^{D}w_t\ell_t$, with weights computed on target-generated training contexts.
DFlash~\citep{chen2026dflash} uses a fixed exponential decay $e^{-(t-1)/\gamma}$ to emphasize early positions (Equation~\eqref{eq:static}). \dpace{}~\citep{wu2026dpace} instead uses the draft's confidence $q_t(y_t^\star)$ as a proxy for $\alpha_t$, smooths it to $\tilde q_t=\rho+(1-\rho)q_t(y_t^\star)$ with $\rho\in(0,1)$, and builds detached weights paired with CE:
\begin{equation}
 w_t^{\rm PACE}=\sum_{k=t}^{D}\prod_{i=1}^{k}\tilde q_i,
 \qquad \mathcal L_{\rm D\text{-}PACE}=\sum_{t=1}^{D} w_t^{\rm PACE}[-\log q_t(y_t^\star)].
\end{equation}
With $\rho>0$, an early low-confidence position cannot zero out all later weights.

\section{\dpalname{}: Acceptance in the Weights, Fit in the Loss}\label{sec:method}\label{sec:dpa-weights}

\subsection{Acceptance-derived weights from overlap}\label{sec:one-token}
\dpace{}'s unsmoothed confidence is overlap with a point mass on the recorded target token: $\sum_v\min\{\delta_{y^\star,v},q_v\}=q(y^\star)$. Standard rejection sampling instead compares the draft with the full target distribution. At a fixed context with target and draft distributions $p_t,q_t$, the verifier accepts $y\sim q_t$ with probability $\min\{1,p_t(y)/q_t(y)\}$. Averaging over $y$ gives the one-token acceptance~\citep{leviathan2023fast,chen2023speculative}
\begin{equation}
 \widehat\alpha_t=\sum_v\min\{p_{t,v},q_{t,v}\}=1-\TV(p_t,q_t),
 \qquad \TV(p,q):=\tfrac12\|p-q\|_1.
 \label{eq:acceptance}
\end{equation}

\dpalname{} applies the weighting rule of Equation~\eqref{eq:relative-credit} to this overlap. On target-generated sequences, the product of overlaps through position $k$ approximates the chance that sampled proposals survive to $k$. As in \dpace{}, we smooth only inside the products, $s_t=\rho+(1-\rho)\widehat\alpha_t$, recompute the weights from the current draft, and detach them:
\begin{equation}
 w_t=\sum_{k=t}^{D}\prod_{i=1}^{k}s_i,
 \qquad \mathcal L_{\rm D\text{-}PAL}=\sum_{t=1}^{D}w_t\,\ell_t,
 \label{eq:weights}
\end{equation}
where $\ell_t$ is the local loss chosen in Section~\ref{sec:dpard-objective}. Equivalently,
\begin{equation}
 w_t=\Bigl(\prod_{i\le t}s_i\Bigr)\Bigl(1+\sum_{k>t}\prod_{t<i\le k}s_i\Bigr),
\end{equation}
the smoothed acceptance through $t$ times the continuation value after $t$ (Figure~\ref{fig:method}).
\input{method_figure}

The SGLang DFlash backend proposes the draft argmax $\hat y_t=\arg\max_v q_{t,v}$ at both temperatures, so proposals are never sampled from $q_t$. Overlap still bounds the acceptance of this proposal.
\begin{proposition}\label{prop:argmax}
(i) Let $\Delta_t$ be the gap between the largest and second-largest target probabilities. If $\TV(p_t,q_t)<\Delta_t/2$, then $\hat y_t=\arg\max_v p_{t,v}$. (ii) $p_t(\hat y_t)\ge\max_v q_{t,v}-\TV(p_t,q_t)$.
\end{proposition}
At $T=0$, the verifier accepts $\hat y_t$ if it matches the target argmax, which (i) guarantees whenever the target margin exceeds $2\TV$. At $T=1$, SGLang's target-only verification accepts it with probability $p_t(\hat y_t)$~\citep{sglang2026dflash}, which (ii) bounds from below (proof in Appendix~\ref{app:argmax}).

\subsection{Log acceptance loss and beyond}\label{sec:dpard-objective}\label{sec:tv-analysis}
\textbf{\dpal{LA}.} Equation~\eqref{eq:relative-credit} pairs each weight with $\nabla\log\alpha_t$, so gradient ascent on $\E[L]$ matches gradient descent on the log-acceptance (LA) loss $\ell_t=-\log\widehat\alpha_t$, which gives \dpalLA{}. For a point-mass target, $\widehat\alpha_t=q_t(y_t^\star)$, so LA becomes CE, as LK Losses notes~\citep{samarin2026lk}, and \dpalLA{} reduces to \dpace{}. At $\rho=0$, gradient descent on \dpalLA{} follows gradient ascent on the overlap-based prefix-sum surrogate (Appendix~\ref{app:position}); as in \dpace{}, we keep the floor $\rho>0$ as a regularizer.

However, LA is not differentiable where a draft probability equals the corresponding target probability. Its gradient is the TV gradient scaled by $1/\widehat\alpha_t$. With respect to the draft logits, the TV gradient is $\tfrac12\,q_t\odot(\sigma_t-\E_{q_t}[\sigma_t])$ with $\sigma_{t,v}=\operatorname{sign}(q_{t,v}-p_{t,v})$~\citep{samarin2026lk}. It does not vanish as $q_t$ approaches $p_t$, and changes sign where $q_{t,v}$ crosses $p_{t,v}$. On the other hand, KL and R\'enyi logit gradients vanish at $q_t=p_t$ (Appendix~\ref{app:local}). Figure~\ref{fig:toy-loss} shows a binary example. Because $w_t$ is detached, any local loss can replace LA without changing the weight formula; in the following, we consider \dpal{KL} and \dpal{R\'enyi} as smooth alternatives to \dpal{LA}.
\input{toy_loss_figure}

\textbf{\dpal{KL}.} Cross-entropy, the local loss of \dpace{}, is smooth but fits one sampled token only. If $y\sim p_t$, then $\E_y[-\log q_t(y)]=H(p_t)+\KL(p_t\Vert q_t)$, and the logit gradient $q_t-e_y$ has mean $q_t-p_t$. Up to the constant $H(p_t)$, CE is thus a one-sample estimate of forward KL, with label noise at every position. Our labels are sampled at positive temperature, so CE also fits a sharper target than the $T=1$ distribution $p_t$ that sets acceptance. \dpalKL{} fits this forward KL directly instead.

\textbf{\dpal{R\'enyi}.} KL fits the full distribution, but it bounds overlap only from below and can be arbitrarily large while overlap is close to one. \dpalRenyi{} instead replaces the minimum in overlap with a geometric mean, the Bhattacharyya coefficient $\mathrm{BC}(p,q)=\sum_v\sqrt{p_vq_v}$, whose scaled negative logarithm is the R\'enyi divergence of order $1/2$~\citep{renyi1961measures}:
\begin{align}
 D_{1/2}(p\Vert q)&=-2\log \mathrm{BC}(p,q),\label{eq:renyi}\\
 \mathcal L_{\text{\dpalRenyi}}&=\sum_{t=1}^{D}w_tD_{1/2}(p_t\Vert q_t).\label{eq:objective}
\end{align}
Its logit gradient is $q_t-r_t$ with $r_{t,v}\propto\sqrt{p_{t,v}q_{t,v}}$ (Appendix~\ref{app:local}), which varies continuously with the target--draft mismatch. The Bhattacharyya coefficient bounds overlap from above and below.

\Needspace{6\baselineskip}
\begin{lemma}[Overlap bounds]
\label{thm:overlap}
For categorical distributions $p,q$, $\alpha(p,q)=1-\TV(p,q)$ satisfies
\begin{equation}
 1-\sqrt{1-e^{-D_{1/2}(p\Vert q)}}\;\le\;\alpha(p,q)\;\le\;e^{-D_{1/2}(p\Vert q)/2}.
 \label{eq:bound}
\end{equation}
\end{lemma}

\Needspace{3\baselineskip}
This bound is the classical inequality between TV and the Bhattacharyya coefficient~\citep{fuchs1999distinguishability} (proof in Appendix~\ref{app:argmax}). Taking negative logarithms gives $D_{1/2}/2\le-\log\widehat\alpha_t\le-\log\bigl(1-\sqrt{1-e^{-D_{1/2}}}\bigr)$. \dpalRenyi{} minimizes the smooth quantity $D_{1/2}$ that controls both bounds, so driving it to zero drives LA to zero. This lower bound on overlap is never below the Bretagnolle--Huber bound from KL~\citep{vanerven2014renyi,canonne2023kltv} (Appendix~\ref{app:kl-bound}).

\subsection{\dpalname{} Algorithm}
Algorithm~\ref{alg:dpard} shows how to compute the \dpalname{} loss. Any of the previously discussed losses (LA, KL, R\'enyi, etc.) can be plugged in for $l_t$.
\input{algorithm}
\FloatBarrier

\section{Experiments}\label{sec:experiments}
\subsection{Experimental setup}
\paragraph{Models and training.}
We train DFlash drafters in SpecForge with frozen Qwen3-4B, Qwen3-8B, and Llama-3.1-8B-Instruct targets~\citep{chen2026dflash,li2026specforge,yang2025qwen3,grattafiori2024llama3}. Within each comparison, methods share target-generated data, optimizer, schedule, and decoding (Appendix~\ref{app:experiments}). Table~\ref{tab:results} uses the Qwen3-4B-Instruct-100K dataset released by the DFlash authors. Following \dpace{}, we use three-layer drafts and block size $B=16$~\citep{chen2026dflash,wu2026dpace}. Request grouping uses these same checkpoints. For the ablations and other targets, we use SpecForge to sample a 36K-example ShareGPT subset; ablations use Qwen3-4B and $B=16$ unless varied.

\paragraph{Evaluation.}
We evaluate on GSM8K~\citep{cobbe2021gsm8k} and MATH-500~\citep{hendrycks2021math,lightman2024verify} for math, HumanEval and MBPP~\citep{chen2021humaneval,austin2021mbpp} for code, and first-turn MT-Bench~\citep{zheng2023mtbench} and Alpaca~\citep{taori2023alpaca} for chat. We run the six-benchmark evaluations with SGLang DFlash on one NVIDIA L40S, one request at a time, at $T=0$ and $T=1$. Our two primary evaluation metrics are speedup ratio (SR) (throughput divided by autoregressive throughput) and acceptance length $\tau$. Six-benchmark means weight benchmarks equally, and every result averages three evaluation seeds. Across seeds, the standard deviation of each mean $\tau$ is 0 at $T=0$ (greedy decoding) and at most 0.02 at $T=1$.

\paragraph{Objectives.}
\staticCE{} and \dpace{} are the baselines. All \dpalname{} objectives share the weight rule of Equation~\eqref{eq:weights} and differ only in the local loss. We compare \dpalLA{}, \dpalKL{}, and \dpalRenyi{} with three losses that contain an explicit TV term: \dpalTV{}, \dpalCETV{}, and \dpalKLTV{} (see Appendix~\ref{app:losses} for details). Table~\ref{tab:results} uses $\rho=0.3$ for \dpalname{}, the best Qwen3-4B value in Table~\ref{tab:smoothing}, and $\rho=0.5$ for \dpace{}, the default that \dpace{} reports as stable over $\rho\in[0.3,0.7]$~\citep{wu2026dpace}. Ablations use $\rho=0.5$ unless varied.

\input{results_table}
\input{grouped_results}
\subsection{\dpalname{} beats \dpace{}}
Table~\ref{tab:results} compares all objectives on Qwen3-4B. \dpalRenyi{} improves on \dpace{} at both temperatures: mean $\tau$ rises from 4.50 to 4.56 at $T=0$ and from 4.20 to 4.28 at $T=1$, and mean SR from 3.11 to 3.14 and from 2.79 to 2.85. The gain holds across benchmarks, since \dpalRenyi{} has the higher $\tau$ on all 12 benchmark--temperature pairs and the higher SR on 10. The other two \dpalname{} objectives, \dpalLA{} and \dpalKL{}, also beat \dpace{} on both metrics at both temperatures. When we batch requests with the same checkpoints, \dpalRenyi{} leads in $\tau$ in all ten settings and in throughput in nine (Table~\ref{tab:grouped}). In the separate 36K ablations at \dpace{}'s own $\rho=0.5$, \dpalRenyi{} still leads \dpace{} by 0.08 in mean $\tau$ at $T=0$ and 0.05 at $T=1$ (Tables~\ref{tab:signal} and~\ref{tab:smoothing}).

\paragraph{LA is not the best local loss.}
Under the same \dpalname{} weights, KL and R\'enyi reach the same or higher mean $\tau$ than LA at both temperatures (Table~\ref{tab:results}). KL leads by 0.01 at $T=0$ and R\'enyi by 0.03 at $T=1$; R\'enyi is our default because it also bounds overlap from above (Lemma~\ref{thm:overlap}).

\paragraph{Pure TV and CE+TV fall short.}\label{sec:local-losses}
\dpalTV{} collapses to $\tau=1.04$, consistent with the small TV gradient at initialization that LK Losses reports~\citep{samarin2026lk}. Adding CE restores training, but \dpalCETV{}, a DSpark- and H-Spec-style mixture, trails \dpalKL{} by 0.18 at $T=0$ and 0.14 at $T=1$ and falls below \dpace{}. Its fixed TV term keeps the nondifferentiable points at $q_{t,v}=p_{t,v}$ (Section~\ref{sec:dpard-objective}), and the toy example in Figure~\ref{fig:conflict-toy} (Appendix~\ref{app:losses}) shows that its CE gradient can raise TV to first order. \dpalKLTV{}, using the adaptive mixture of losses proposed in LK Losses, shifts weight to KL at low-acceptance positions and stays within 0.04 of \dpalLA{}, \dpalKL{}, and \dpalRenyi{}.

\subsection{Ablations separate the loss from the weights}\label{sec:ablations}
\paragraph{Distribution fitting beats cross-entropy.}
Table~\ref{tab:signal} changes only the local loss, from CE to R\'enyi, under each weight signal. The switch raises mean $\tau$ by 0.08 at $T=0$ and 0.06 at $T=1$ under \dpace{}'s confidence weights ($\rho=0.5$), and by 0.09 and 0.07 under overlap weights ($\rho=0.3$). With R\'enyi at the same floor $\rho=0.5$, the two weight signals give similar mean $\tau$ (3.58 for both at $T=0$; 3.42 and 3.41 at $T=1$; Tables~\ref{tab:signal} and~\ref{tab:smoothing}). We keep overlap as the weight signal because it matches rejection-sampling acceptance, reduces to \dpace{}'s weights for a point-mass target, and bounds argmax acceptance (Proposition~\ref{prop:argmax}).

\paragraph{\dpalname{} weights beat fixed decay on baseline losses.}\label{sec:position-ablation}
In Table~\ref{tab:dynamic-controls}, we compare DFlash's static weight decay against \dpalname{}'s weight schedule across four losses from the literature: CE, used in \citet{chen2026dflash}; LA and KL+TV, used in \citet{samarin2026lk}; and CE+TV, used in \citet{cheng2026dspark}. At $T=0$, \dpalname{} weights raise mean $\tau$ over DFlash's schedule by 0.26 for CE, 0.29 for LA, 0.33 for KL+TV, and 0.08 for CE+TV. Gains are smaller at $T=1$ but hold for every loss.

\begin{table}[t]

\begin{minipage}[t]{0.485\linewidth}
\vspace{0pt}
\input{signal_results}
\end{minipage}\hfill
\begin{minipage}[t]{0.485\linewidth}
\vspace{0pt}
\input{dynamic_control_results}
\end{minipage}
\end{table}

\paragraph{\dpalRenyi{} leads at every block size and depth.}
Larger blocks help up to $B=12$: the mean $\tau$ of \dpalRenyi{} rises from $B=8$ to $B=12$ and levels off at $B=16$ (Table~\ref{tab:block-size}). Deeper drafts help both methods, and with five layers \dpalRenyi{} still leads \dpace{} by 0.06 at $T=0$ and 0.07 at $T=1$ (Table~\ref{tab:draft-depth}).

\input{capacity_results}

\paragraph{The floor prevents collapse.}
Without the floor ($\rho=0$), mean $\tau$ falls to $1.03$--$1.60$, likely because an early low-overlap position drives every later prefix product toward zero (Table~\ref{tab:smoothing}). Any floor in $\{0.3,0.5\}$ works: the two stay within 0.04 of each other on every configuration.

\subsection{Gains hold on other targets and inference settings}\label{sec:grouped}
\paragraph{Gains transfer to 8B targets.}
\dpalRenyi{} has the highest mean SR and $\tau$ on both 8B targets at both temperatures (Table~\ref{tab:targets}). The gain is largest on Llama, where \dpalRenyi{} leads \dpace{} by 0.13 at $T=0$ and 0.12 at $T=1$, while \dpace{} falls below \staticCETV{}.

\begin{table}[t]

\begin{minipage}[t]{0.485\linewidth}
\vspace{0pt}
\input{smoothing_results}
\end{minipage}\hfill
\begin{minipage}[t]{0.485\linewidth}
\vspace{0pt}
\input{target_results}
\end{minipage}
\end{table}

\input{thinking_results}
\paragraph{Gains hold in thinking mode.}\label{sec:thinking}
With thinking enabled, all three \dpalname{} losses match or improve \dpace{} in mean $\tau$; \dpalRenyi{} leads at $T=0$ and \dpalKL{} at $T=1$ (Table~\ref{tab:thinking}).

\section{Related Work}
\paragraph{Parallel drafters.}
Medusa and DFlash propose every block position in parallel~\citep{cai2024medusa,chen2026dflash}. Their per-position marginals ignore tokens drafted earlier in the block, which LiLiCorr calls the marginals problem~\citep{rusanovsky2026lilicorr}. EAGLE, Hydra, Domino, and DSpark condition later proposals on earlier tokens~\citep{li2024eagle,li2025eagle3,ankner2024hydra,huang2026domino,cheng2026dspark}. LiLiCorr reranks adjacent top-$k$ candidates with a head co-trained with DFlash, DFlash2 adds a predecessor-conditioned selector, and DFlow carries verifier representations between rounds~\citep{rusanovsky2026lilicorr,inco2026dflash2,zhang2026dflow}. H-Spec replaces the drafter-side KV cache with a hybrid Mamba--attention drafter trained with CE+TV and fixed decay~\citep{jiang2026hspec}. These methods change what the drafter conditions on; orthogonally, \dpalname{} changes the objective that trains the marginals and can be combined with them.

\paragraph{Acceptance-aware position weighting.}
Several methods adapt position weights to acceptance. \dpace{} weights CE by a surrogate of expected length built from draft confidence~\citep{wu2026dpace}, following EAGLE-2's observation that draft confidence tracks acceptance~\citep{li2024eagle2}. PARD-2 instead uses the product of the target's probabilities of the preceding recorded tokens~\citep{an2026pard2}. Other methods tie supervision to where the draft first fails: GRIFFIN propagates top-$k$ alignment masks~\citep{hu2025griffin}, Spec-AUF keeps CE through the first predicted failure~\citep{yang2026specauf}, and \citet{whalen2026teaching} study first-error supervision with an expected-length chain reward. VAT keeps full weight up to the first simulated rejection, restarts the decay there, and trains a verification head~\citep{gu2026vat}. It also finds that weighting by the exact derivative of expected length, computed from confidence, trails \dpace{}'s weights. \dpalname{} weights keep \dpace{}'s weighting rule and floor and use overlap, $1-\TV(p,q)$, in place of confidence.

\Needspace{10\baselineskip}
\paragraph{Local losses for acceptance.}
Other works change the local loss. DistillSpec finds that the best divergence depends on the task and decoding strategy~\citep{zhou2024distillspec}, and DART fits target-to-draft KL with fixed decay~\citep{liu2026dart}. Draft-OPD and AdaFlash train on draft-generated contexts with forward and reverse KL or clipped reverse KL~\citep{lei2026draftopd,qian2026adaflash}. Closer to our setting, Bebop optimizes a normalized expected length built from TV acceptance~\citep{li2026bebop}. LK Losses minimizes the negative log acceptance, the same term as our LA, or an adaptive KL--TV mixture, and notes that the negative log acceptance reduces to CE for a point-mass target~\citep{samarin2026lk}. It keeps fixed position decay and leaves learnable or data-dependent weighting across draft heads to future work. DSpark and Uno mix CE or distillation with TV terms~\citep{cheng2026dspark,sahoo2026uno}. \dpalname{} puts acceptance into the weights, so the loss can be a smooth distributional fit (Section~\ref{sec:experiments}).

\Needspace{7\baselineskip}
\section{Conclusion}
We derived \dpalname{} from the expected accepted length in self-speculative decoding by replacing the point-mass confidence from \dpace{}~\citep{wu2026dpace} with the full-distribution overlap. The derivation gives acceptance-derived position weights with log acceptance as the local loss. Furthermore, we show that other losses, such as KL and R\'enyi-$1/2$, match or exceed the derived log acceptance in average acceptance, while enjoying theoretical guarantees and smoothness, and show they can be combined with the detached \dpalname{} weight schedule. Experiments across six benchmarks show up to 5\% longer acceptance lengths and higher throughput than \dpace{}, and up to 11\% higher than DFlash~\citep{chen2026dflash}.

\paragraph{Limitations and future work.}
\dpalname{} weights assume that proposals within a block do not condition on each other, as in DFlash; drafters such as DSpark or EAGLE would need weights that model this dependence~\citep{cheng2026dspark,li2024eagle,li2025eagle3}. \dpalname{} weights could also train the parallel logits of DFlash2 or LiLiCorr~\citep{inco2026dflash2,rusanovsky2026lilicorr}.
\label{sec:main-end}

\section*{Reproducibility Statement}
Appendix~\ref{app:experiments} gives training and evaluation settings. Sections~\ref{sec:position-training}--\ref{sec:method} and Algorithm~\ref{alg:dpard} specify the objectives, with supporting derivations in Appendices~\ref{app:local} and~\ref{app:position}. We provide the training code at \url{https://github.com/Lucas-TY/D-PAL-SpecForge}.
\Needspace{4\baselineskip}
\section*{AI Use Statement}
Generative AI tools assisted with drafting and revising text. Target language models generated the training responses as described in Section~\ref{sec:experiments}. The authors are responsible for the final manuscript.
\bibliographystyle{plainnat}
\bibliography{references}
\clearpage
\appendix
\input{experiment_details}

\section{Additional Derivations}
\subsection{R\'enyi fitting gradient}\label{app:local}
Fix $p$ and let $q=\operatorname{softmax}(z)$. Using $\partial q_j/\partial z_v=q_j(\mathbf1_{j=v}-q_v)$,
\[
 \frac{\partial \mathrm{BC}(p,q)}{\partial z_v}
 =\frac12\bigl(\sqrt{p_vq_v}-q_v\mathrm{BC}(p,q)\bigr),\qquad
 \frac{\partial D_{1/2}}{\partial z_v}
 =-\frac{2}{\mathrm{BC}(p,q)}\frac{\partial \mathrm{BC}(p,q)}{\partial z_v}=q_v-r_v.
\]
Here $r_v=\sqrt{p_vq_v}/\mathrm{BC}(p,q)$ is the normalized geometric mean.

\Needspace{8\baselineskip}
\paragraph{Comparison with KL.}\label{app:kl-bound}
R\'enyi divergence is nondecreasing in order, so $D_{1/2}(p\Vert q)\le\KL(p\Vert q)$~\citep{vanerven2014renyi}. Bretagnolle--Huber gives $\alpha(p,q)\ge1-\sqrt{1-e^{-\KL(p\Vert q)}}$~\citep{canonne2023kltv}. For the same $p,q$, Equation~\eqref{eq:bound} gives an overlap lower bound at least as high as the KL bound.

\subsection{Proofs of Proposition~\ref{prop:argmax} and Lemma~\ref{thm:overlap}}\label{app:argmax}
\paragraph{Proof of Lemma~\ref{thm:overlap}.}
The upper bound follows from $\min\{p_j,q_j\}\le\sqrt{p_jq_j}$. For the lower bound, set $u_j=\sqrt{p_j}$ and $v_j=\sqrt{q_j}$. Both have unit norm, and $u^\top v=\mathrm{BC}(p,q)$:
\[
\begin{aligned}
 2\TV(p,q)&=\sum_j |(u_j-v_j)(u_j+v_j)|\\
 &\le \|u-v\|_2\,\|u+v\|_2
 =2\sqrt{1-\mathrm{BC}(p,q)^2}
 =2\sqrt{1-e^{-D_{1/2}(p\Vert q)}}.
\end{aligned}
\]
Hence $\alpha(p,q)=1-\TV(p,q)\ge1-\sqrt{1-e^{-D_{1/2}(p\Vert q)}}$.

\paragraph{Proof of Proposition~\ref{prop:argmax}.}
Write $p=p_t$, $q=q_t$, $\Delta=\Delta_t$, $\hat y=\arg\max_v q_v$, and $v^\star=\arg\max_v p_v$. (i) Suppose $\hat y=j\ne v^\star$, so $q_j\ge q_{v^\star}$. Then
\[
 2\TV(p,q)\ge(p_{v^\star}-q_{v^\star})+(q_j-p_j)=(p_{v^\star}-p_j)+(q_j-q_{v^\star})\ge \Delta,
\]
which contradicts $\TV(p,q)<\Delta/2$. (ii) Since $\TV(p,q)=\sum_v(q_v-p_v)_+\ge q_{\hat y}-p_{\hat y}$, we have $p_{\hat y}\ge\max_v q_v-\TV(p,q)$.

\Needspace{5\baselineskip}
\subsection{TV gradients and position weights}\label{app:position}\label{app:smoothing}
\paragraph{TV coefficients.}
Fix the target model and target-generated training contexts. Differentiating the prefix sum of training overlaps $\widehat\alpha_t=1-\TV(p_t,q_t)$ gives
\begin{equation}
 \nabla_\theta\sum_{k=1}^{D}\prod_{i=1}^{k}\widehat\alpha_i
 =\sum_{t=1}^{D}\left[
 \sum_{k=t}^{D}\prod_{\substack{1\le i\le k\\i\ne t}}\widehat\alpha_i
 \right]\nabla_\theta\widehat\alpha_t.
 \label{eq:tv-credit}
\end{equation}
Multiplying this coefficient by $\widehat\alpha_t$ gives the relative-contribution form in Equation~\eqref{eq:relative-credit}, evaluated at training overlaps.

\paragraph{Gradient of the smoothed prefix sum.}
For positive $s_t=\rho+(1-\rho)\widehat\alpha_t$, direct differentiation of the smoothed prefix sum gives
\[
 \nabla_\theta\sum_{k=1}^{D}\prod_{i=1}^{k}s_i
 =\sum_{t=1}^{D}\frac{1-\rho}{s_t}w_t\,\nabla_\theta\widehat\alpha_t.
\]
At $\rho=0$, $s_t=\widehat\alpha_t$ and gradient descent on \dpalLA{} follows this direction; for $\rho>0$, the factor $(1-\rho)/s_t$ replaces $1/\widehat\alpha_t$, and we treat the floor as a regularizer, as \dpace{} does.

\end{document}

%% file: preamble.tex
\usepackage[letterpaper,textwidth=5.5in,textheight=9in,centering]{geometry}
\usepackage{times}
\usepackage[round,authoryear]{natbib}
\usepackage{amsmath,amssymb,amsthm,booktabs}
\usepackage{threeparttable}
\usepackage{array}
\usepackage{graphicx}
\usepackage{wrapfig}
\usepackage{tikz}
\usepackage{float}
\usepackage{placeins}
\usepackage{needspace}

\usetikzlibrary{arrows.meta}
\usepackage[T1]{fontenc}
\usepackage{xurl}
\usepackage{hyperref}
\hypersetup{
  colorlinks=true,
  linkcolor={red},
  citecolor={orange!70!black},
  urlcolor={blue!80!black},
  pdftitle={Beyond the Point Mass: Acceptance-Derived Weights and Distributional Losses for Speculative Decoding},
  pdfauthor={Tianyu Wu; Zhuohan Wang; Jonathan Geuter; Yu Yao; Weifan Jiang; Chengxi Zhang; Ju Li; Minlan Yu},
  pdfsubject={Speculative decoding},
  pdfkeywords={D-PAL, speculative decoding, draft model training},
}
\newcommand{\TV}{\operatorname{TV}}
\newcommand{\KL}{\operatorname{KL}}

\newcommand{\E}{\mathbb E}

\newtheorem{proposition}{Proposition}
\newtheorem{lemma}{Lemma}

\newcounter{algorithm}
\date{}
\usepackage{pgfplots}
\pgfplotsset{compat=1.18}
\newcommand{\dpace}{D-PACE}
\newcommand{\dpalname}{D-PAL}
\newcommand{\dpal}[1]{\mbox{\dpalname[#1]}}
\newcommand{\dpalRenyi}{\dpal{R\'enyi}}
\newcommand{\dpalKL}{\dpal{KL}}
\newcommand{\dpalCE}{\dpal{CE}}
\newcommand{\dpalLA}{\dpal{LA}}
\newcommand{\dpalTV}{\dpal{TV}}
\newcommand{\dpalCETV}{\dpal{CE+TV}}
\newcommand{\dpalKLTV}{\dpal{KL+TV}}
\newcommand{\staticname}{Static}
\newcommand{\staticx}[1]{\mbox{\staticname[#1]}}
\newcommand{\staticCE}{DFlash}
\newcommand{\staticCETV}{\staticx{CE+TV}}

%% file: method_figure.tex
\begingroup
\setlength{\intextsep}{3pt}
\ifdefined\ChineseDraft
\begin{figure}[H]
\else
\begin{figure}[!htbp]
\fi
\centering
\definecolor{creditBlueBg}{HTML}{DDEDF8}
\definecolor{creditGreenBg}{HTML}{E1EDDD}
\definecolor{creditBlue}{HTML}{A7CDE5}
\definecolor{creditGreen}{HTML}{AFD7B2}
\definecolor{lossPeach}{HTML}{FFF4E8}
\definecolor{targetBar}{HTML}{DF9E6A}
\definecolor{draftBar}{HTML}{80B1CF}
\definecolor{figureInk}{HTML}{343434}
\begin{tikzpicture}[x=1cm,y=1cm,
 every node/.style={font=\small,text=figureInk,align=center},
 compact/.style={font=\scriptsize},
 every path/.style={draw=figureInk,line width=.6pt}]
\path[use as bounding box,draw=none] (0,0.55) rectangle (13.7,6.85);
\ifdefined\ChineseDraft
 \def\AccuHeading{接受至位置 $t$}
 \def\LocalHeading{当前位置 $t$}
 \def\ContHeading{接受位置 $t$ 后的延伸价值}
 \def\PointLabel{单 token 对齐}
 \def\DistLabel{分布对齐}
\else
 \def\AccuHeading{Acceptance through $t$}
 \def\LocalHeading{Local loss at $t$}
 \def\ContHeading{Continuation after $t$}
 \def\PointLabel{Single-token fitting}
 \def\DistLabel{Distribution fitting}
\fi
\ifdefined\ChineseDraft
 \def\ConfidenceRow{D-PACE}
 \def\DistributionRow{\dpalname{} losses}
 \def\RowTitleFont{\normalsize\bfseries}
\else
 \def\ConfidenceRow{\dpace{} (confidence)}
 \def\DistributionRow{\dpalname{} (overlap, ours)}
 \def\RowTitleFont{\small\bfseries}
\fi
\node[font=\small\bfseries] at (2.10,6.66) {\AccuHeading};
\node[font=\small\bfseries] at (6.65,6.66) {\LocalHeading};
\node[font=\small\bfseries] at (11.40,6.66) {\ContHeading};
\node[font=\large] at (4.30,6.66) {$\times$};
\node[font=\large] at (9.00,6.66) {$\times$};
\foreach \shift/\method/\bg/\nodeFill/\targetlabel/\signal/\acceptance in {
 3.00/{\ConfidenceRow}/creditBlueBg/creditBlue/\PointLabel/{\tilde q_{\pos}}/{q_i(y_i^\star)},
 0/{\DistributionRow}/creditGreenBg/creditGreen/\DistLabel/{s_{\pos}}/{\widehat\alpha_i=1-\mathrm{TV}(p_i,q_i)}
}{
 \begin{scope}[yshift=\shift cm]
 \path[fill=\bg,draw=none,rounded corners=5pt] (0,.62) rectangle (13.7,3.45);
 \node[anchor=west,font=\RowTitleFont] at (.25,3.18) {\method};
 \foreach \xx/\pos in {.70/1,2.05/2,3.65/t,9.68/{t+1},11.28/{t+2},12.95/D}{
  \node[compact,draw,fill=\nodeFill,rounded corners=2pt,minimum width=1.18cm,minimum height=.52cm,inner xsep=2pt] at (\xx,1.80) {$\signal$};
  \node[compact] at (\xx,1.28) {$\pos$};
 }
 \node[compact] at (2.85,1.80) {$\cdots$};
 \node[compact] at (12.16,1.80) {$\cdots$};
 \path[fill=lossPeach,rounded corners=4pt] (4.45,.72) rectangle (8.85,3.22);
 \node[text height=1.5ex,text depth=.5ex] at (6.65,2.99) {\targetlabel};
 \node[compact,text height=1.5ex,text depth=.5ex] at (2.10,2.60) {\ifdefined\ChineseDraft 训练信号\else Weight signal\fi: $\acceptance$};
 \path[draw=none,fill=targetBar] (4.90,2.56) rectangle (5.10,2.64);
 \node[anchor=west,compact,inner sep=0pt] at (5.16,2.60) {Target $p_t$};
 \path[draw=none,fill=draftBar] (6.78,2.56) rectangle (6.98,2.64);
 \node[anchor=west,compact,inner sep=0pt] at (7.04,2.60) {Draft $q_t$};
 \end{scope}
}
\foreach \shift in {0,3.00}{
 \begin{scope}[yshift=\shift cm]
 \ifdim\shift pt>0pt
  \path[fill=white,draw=targetBar!65!figureInk,line width=.5pt,rounded corners=2pt]
    (5.17,1.68) rectangle (8.57,2.04);
 \else
  \path[fill=white,draw=targetBar!65!figureInk,line width=.5pt,rounded corners=2pt]
    (5.17,.97) rectangle (8.57,2.40);
 \fi
 \draw[draw=figureInk!30,line width=.4pt] (5.35,1.02)--(5.35,2.35);
 \foreach \yy/\teacher/\student in {2.22/.12/.30,1.86/.52/.22,1.50/.26/.16,1.14/.10/.32}{
  \def\barOpacity{1}
  \ifdim\shift pt>0pt
   \ifdim\yy pt=1.86pt\else\def\barOpacity{.24}\fi
  \fi
  \draw[draw=figureInk!10,line width=.35pt] (5.35,\yy)--(8.35,\yy);
  \path[draw=none,fill=targetBar,opacity=\barOpacity]
    (5.35,\yy+.025) rectangle ({5.35+3*\teacher},\yy+.105);
  \path[draw=none,fill=draftBar]
    (5.35,\yy-.080) rectangle ({5.35+3*\student},\yy);
 }
 \node[compact,text=figureInk!65] at (5.35,.87) {$0$};
 \node[compact,text=figureInk!65] at (8.35,.87) {$1$};
 \ifdim\shift pt>0pt
  \node[compact] at (4.91,1.86) {$y_t^\star$};
 \fi
 \end{scope}
}
\end{tikzpicture}
\ifdefined\ChineseDraft
\caption{位置加权与局部拟合。\dpace{} 使用目标 token 置信度，\dpalname{} 损失使用重叠。累计项包含 $t$；延伸价值计入 $t$ 及后续延伸。两组使用相同的分布：上排白框标出 CE 监督的目标 token，下排白框涵盖完整目标与草稿分布。淡化横条不表示其 logit 梯度为零。}
\else
\caption{\textbf{\dpalname{} uses the full target distribution for position weighting and, with KL or R\'enyi, for local fitting.} Top: \dpace{} weights by confidence and fits one token with CE; faded bars are target mass CE ignores. Bottom: \dpalname{} weights by overlap and, with a distributional loss such as KL or R\'enyi, fits $q_t$ to all of $p_t$. Each weighted term $w_t\ell_t$ is the product of the acceptance through $t$ (left), the local loss at $t$ (center), and the continuation value after $t$ (right).}
\fi
\label{fig:method}
\end{figure}
\endgroup

%% file: toy_loss_figure.tex
\ifdefined\ChineseDraft
\begin{figure}[!htbp]
\centering
\includegraphics[width=\linewidth]{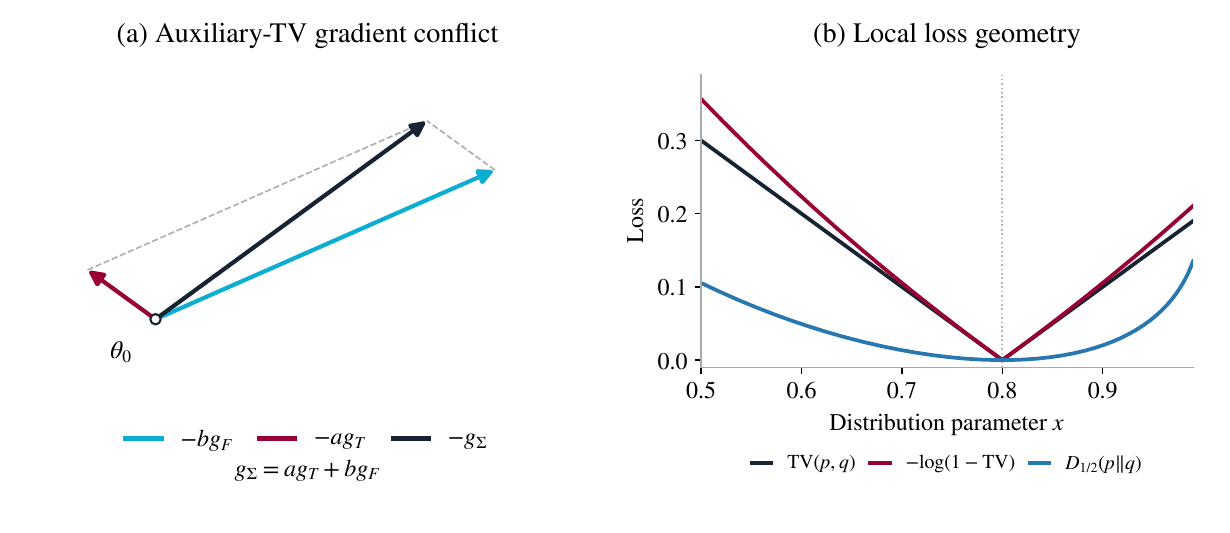}
\caption{梯度干扰与非光滑性的 toy 示例。(a) 同一参数点的负加权梯度：融合损失 $a\TV+b\mathcal L_F$ 沿自身梯度下降，TV 则在一阶上上升（式~\eqref{eq:conflict}；具体损失与系数见附录~\ref{app:conflict}）。(b) 对二元分布 $p=(0.8,0.2)$、$q(x)=(x,1-x)$，TV 在 $x=0.8$ 处有尖点，而 Rényi 损失 $D_{1/2}$ 平滑。}
\label{fig:toy-loss}
\end{figure}
\else
\begin{figure}[!t]
\centering
\begin{minipage}[c]{0.50\linewidth}
\centering
\begin{tikzpicture}
\begin{axis}[width=\linewidth,height=4.0cm,xmin=0.25,xmax=0.75,ymin=0,ymax=0.31,
 axis lines=left,axis line style={-,gray!70},tick style={gray!70},
 xtick={0.25,0.5,0.75},ytick={0,0.1,0.2,0.3},tick label style={font=\scriptsize},
 xlabel={Draft parameter $x$},ylabel={Local loss},label style={font=\scriptsize},
 legend style={font=\scriptsize,at={(0.5,1.0)},anchor=north,legend columns=3,draw=none,fill=none},
 legend cell align=left,
 ymajorgrids,grid style={gray!15},samples=121,domain=0.25:0.75,no markers]
\addplot[line width=1.0pt,color={rgb,255:red,196;green,78;blue,82}] {-ln(1-abs(x-0.5))};
\addplot[line width=1.0pt,color=gray!75!black] {0.5*ln(0.5/x)+0.5*ln(0.5/(1-x))};
\addplot[line width=1.8pt,color={rgb,255:red,31;green,119;blue,180}] {-2*ln(sqrt(0.5*x)+sqrt(0.5*(1-x)))};
\draw[dashed,gray!60] (axis cs:0.5,0) -- (axis cs:0.5,0.24);
\node[anchor=west,font=\scriptsize,text=gray!80!black] at (axis cs:0.505,0.14) {$q=p$};
\legend{LA,KL,R\'enyi (default)}
\end{axis}
\end{tikzpicture}
\end{minipage}\hfill
\begin{minipage}[c]{0.46\linewidth}
\caption{\textbf{Near $q=p$, LA's slope stays near $\pm1$ and flips sign, while KL and R\'enyi-$1/2$ flatten out.} Binary example with target $p=(0.5,0.5)$ and draft $q=(x,1-x)$; slopes are with respect to $x$, and all three losses are zero at $x=0.5$.}
\label{fig:toy-loss}
\end{minipage}
\end{figure}
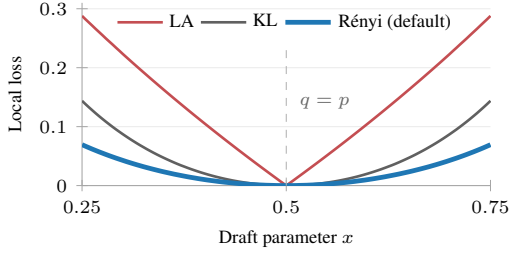
\fi

%% file: algorithm.tex
\begingroup
\setlength{\intextsep}{4pt}
\begin{figure}[H]
\refstepcounter{algorithm}\label{alg:dpard}
\centering
\begin{minipage}{.95\linewidth}
\hrule\smallskip
\textbf{\ifdefined\ChineseDraft 算法 1：\dpalname{} 损失计算\else Algorithm 1: \dpalname{} loss computation\fi}
\smallskip\hrule\smallskip
\small
\ifdefined\ChineseDraft
\textbf{输入：} 训练时的固定归一化目标分布 $p$、归一化草稿分布 $q$、有效位置掩码 $m$、平滑系数 $\rho$、目标 D-PALA 或 \dpalRenyi{}。
\else
\textbf{Input:} target $p$, draft $q$, valid mask $m$, smoothing $\rho$, and chosen local loss $\ell_t$.
\fi
\begin{tabbing}
\quad\=\quad\=\kill
\textbf{for} each nonempty valid block $b$:\\[1pt]
\> \ifdefined\ChineseDraft 按原顺序选取 $m_{b,t}=1$ 的候选，重编号为 $t=1,\ldots,D_b$。\else Select proposals with $m_{b,t}=1$ in order; index them by $t=1,\ldots,D_b$.\fi\\[1pt]
\> $\widehat\alpha_t\gets\sum_v\min(p_{t,v},q_{t,v})$;\quad $s_t\gets\rho+(1-\rho)\widehat\alpha_t$ \quad\textbf{for all} $t$\\[1pt]
\> $\widetilde R_0\gets1$;\quad $w_{D_b+1}\gets0$\\[1pt]
\> \textbf{for} $t=1,\ldots,D_b$: \quad $\widetilde R_t\gets s_t\widetilde R_{t-1}$\ifdefined\ChineseDraft\else\quad\(\triangleright\) \textit{Smoothed prefix product}\fi\\[1pt]
\> \textbf{for} $t=D_b,\ldots,1$: \quad $w_t\gets w_{t+1}+\widetilde R_t$\ifdefined\ChineseDraft\else\quad\(\triangleright\) \textit{Position weights}\fi\\[1pt]
\> \textbf{Detach} $w_{1:D_b}$.\\[1pt]
\ifdefined\ChineseDraft
\> \textbf{if} D-PALA: $\ell_t\gets-\log\widehat\alpha_t$ \quad\textbf{for all} $t$\\[1pt]
\> \textbf{if} \dpalRenyi{}: $\ell_t\gets-2\log\sum_v\sqrt{p_{t,v}q_{t,v}}$ \quad\textbf{for all} $t$\\[1pt]
\fi
\> $\mathcal L_b\gets\sum_{t=1}^{D_b}w_t\ell_t$\\[1pt]
\textbf{return} \ifdefined\ChineseDraft 每条序列内的有效块均值，再对序列取平均。\else Mean of $\mathcal L_b$ within each sequence, then across sequences.\fi
\end{tabbing}
\ifdefined\ChineseDraft
两次权重扫描耗时 $O(D_b)$，词表求和耗时 $O(D_bV)$。Rényi 损失在 FP32 中由 log-softmax 与 logsumexp 计算。
\fi
\smallskip\hrule
\end{minipage}
\end{figure}
\endgroup

%% file: results_table.tex
\ifdefined\ChineseDraft
\begin{table}[t]
\else
\begin{table}[t]
\fi
\vspace{-20pt}
\centering
\small
\setlength{\tabcolsep}{3pt}
\renewcommand{\arraystretch}{1.12}
\begin{threeparttable}
\ifdefined\ChineseDraft
\caption{\textbf{主要结果。}Qwen3-4B、三层 DFlash、$B=16$ 下相对自回归解码的加速比（SR）与接受长度 $\tau$。\dpalname{} (loss) 为我们构造的重叠加权损失对照；\dpalRenyi{} 使用 Rényi。LK 指 KL--TV 混合损失。Mean 根据未舍入的六基准结果等权计算。}
\else
\caption{\textbf{\dpalRenyi{}, \dpalLA{}, and \dpalKL{} beat \dpace{}; \dpalTV{} and \dpalCETV{} fall short.} Qwen3-4B target, three-layer DFlash drafts, $B=16$, 100K training examples, $\rho=0.3$ for \dpalname{} and $0.5$ for \dpace{}. SR is speedup over autoregressive decoding, $\tau$ is acceptance length, and bold marks the best value.}
\fi
\label{tab:results}
\begin{tabular*}{\linewidth}{@{}l@{\extracolsep{\fill}\hspace{3pt}}*{5}{w{r}{19pt}@{\extracolsep{0pt}\hspace{1.5pt}}w{r}{19pt}@{\extracolsep{\fill}\hspace{9pt}}}w{r}{19pt}@{\extracolsep{0pt}\hspace{1.5pt}}w{r}{19pt}@{\extracolsep{\fill}\hspace{9pt}}w{r}{19pt}@{\extracolsep{0pt}\hspace{1.5pt}}w{r}{19pt}@{}}
\toprule
& \multicolumn{4}{c}{\textsc{Math}} & \multicolumn{4}{c}{\textsc{Code}} & \multicolumn{4}{c}{\textsc{Chat}} & \multicolumn{2}{c}{}\\
\cmidrule(lr){2-5}\cmidrule(lr){6-9}\cmidrule(lr){10-13}
& \multicolumn{2}{c}{\makebox[0pt]{GSM8K}} & \multicolumn{2}{c}{\makebox[0pt]{\hspace{-2pt}MATH-500}} & \multicolumn{2}{c}{\makebox[0pt]{\hspace{2pt}HumanEval}} & \multicolumn{2}{c}{\makebox[0pt]{MBPP}} & \multicolumn{2}{c}{\makebox[0pt]{MT-Bench}} & \multicolumn{2}{c}{\makebox[0pt]{Alpaca}} & \multicolumn{2}{c}{\makebox[0pt]{\textbf{Mean}}}\\
\cmidrule(lr){2-3}\cmidrule(lr){4-5}\cmidrule(lr){6-7}\cmidrule(lr){8-9}\cmidrule(lr){10-11}\cmidrule(lr){12-13}\cmidrule(l){14-15}
Method & SR & $\tau$ & SR & $\tau$ & SR & $\tau$ & SR & $\tau$ & SR & $\tau$ & SR & $\tau$ & SR & $\tau$\\
\midrule
\textit{Baselines} & \multicolumn{14}{c}{\textit{Temperature $T=0$}}\\
\midrule
\staticCE{} & 3.20 & 4.50 & 3.94 & 5.46 & 3.33 & 4.67 & 3.06 & 4.28 & 1.97 & 3.15 & 1.77 & 2.71 & 2.88 & 4.13\\ 
\dpace{} & 3.50 & 4.96 & 4.35 & 6.06 & 3.60 & 5.09 & 3.34 & 4.65 & 2.05 & 3.36 & 1.83 & 2.87 & 3.11 & 4.50\\ 
\cmidrule(lr){1-15}
\multicolumn{15}{l}{\textit{\dpalname{} weights}}\\
\dpalLA{} & 3.50 & 5.03 & 4.32 & 6.06 & 3.65 & 5.15 & \textbf{3.38} & 4.70 & 2.04 & 3.41 & 1.84 & 2.90 & 3.12 & 4.54\\ 
\dpalKL{} & 3.50 & 5.06 & \textbf{4.43} & \textbf{6.11} & \textbf{3.68} & \textbf{5.19} & \textbf{3.38} & \textbf{4.71} & \textbf{2.08} & 3.42 & \textbf{1.87} & \textbf{2.92} & \textbf{3.16} & \textbf{4.57}\\ 
\dpalRenyi{} & \textbf{3.53} & \textbf{5.07} & 4.39 & 6.10 & 3.66 & 5.17 & 3.31 & 4.68 & \textbf{2.08} & \textbf{3.44} & 1.86 & \textbf{2.92} & 3.14 & 4.56\\ 
\cmidrule(lr){1-15}
\dpalTV{} & 0.75 & 1.01 & 0.76 & 1.02 & 0.77 & 1.04 & 0.77 & 1.03 & 0.77 & 1.05 & 0.77 & 1.07 & 0.77 & 1.04\\ 
\dpalCETV{} & 3.39 & 4.89 & 4.24 & 5.86 & 3.50 & 4.92 & 3.27 & 4.54 & 1.94 & 3.23 & 1.77 & 2.86 & 3.02 & 4.39\\ 
\dpalKLTV{} & 3.49 & 5.03 & 4.33 & 6.05 & 3.64 & 5.14 & 3.33 & 4.68 & 2.04 & 3.40 & 1.85 & 2.88 & 3.11 & 4.53\\ 
\midrule
\textit{Baselines} & \multicolumn{14}{c}{\textit{Temperature $T=1$}}\\
\midrule
\staticCE{} & 2.99 & 4.29 & 3.25 & 4.80 & 3.07 & 4.40 & 2.85 & 4.09 & 1.84 & 3.04 & 1.71 & 2.71 & 2.62 & 3.89\\ 
\dpace{} & 3.24 & 4.76 & 3.54 & 5.31 & 3.30 & 4.75 & 3.03 & 4.41 & 1.90 & 3.18 & 1.75 & 2.82 & 2.79 & 4.20\\ 
\cmidrule(lr){1-15}
\multicolumn{15}{l}{\textit{\dpalname{} weights}}\\
\dpalLA{} & 3.28 & 4.78 & 3.52 & 5.26 & \textbf{3.37} & 4.81 & 3.05 & 4.48 & 1.91 & 3.26 & 1.76 & 2.85 & 2.82 & 4.24\\ 
\dpalKL{} & 3.28 & 4.81 & 3.49 & 5.27 & 3.36 & 4.82 & 3.05 & 4.47 & 1.91 & 3.25 & 1.78 & 2.85 & 2.81 & 4.25\\ 
\dpalRenyi{} & \textbf{3.34} & \textbf{4.84} & 3.52 & \textbf{5.35} & \textbf{3.37} & 4.82 & \textbf{3.10} & 4.48 & \textbf{1.94} & 3.28 & \textbf{1.81} & \textbf{2.89} & \textbf{2.85} & \textbf{4.28}\\ 
\cmidrule(lr){1-15}
\dpalTV{} & 0.75 & 1.01 & 0.76 & 1.02 & 0.77 & 1.04 & 0.77 & 1.03 & 0.77 & 1.05 & 0.77 & 1.06 & 0.76 & 1.04\\ 
\dpalCETV{} & 3.18 & 4.64 & 3.48 & 5.19 & 3.23 & 4.61 & 2.96 & 4.32 & 1.85 & 3.11 & 1.69 & 2.78 & 2.73 & 4.11\\ 
\dpalKLTV{} & 3.25 & 4.77 & \textbf{3.58} & 5.34 & \textbf{3.37} & \textbf{4.88} & 3.07 & \textbf{4.49} & 1.91 & \textbf{3.30} & 1.77 & 2.84 & 2.83 & 4.27\\ 
\bottomrule
\end{tabular*}
\end{threeparttable}
\end{table}

%% file: grouped_results.tex
\begin{table}[t]
\vspace{-20pt}
\centering
\small
\setlength{\tabcolsep}{3pt}
\renewcommand{\arraystretch}{1.1}
\ifdefined\ChineseDraft
\caption{128条固定 MATH-500 提示上的分组请求评测。目标为 Qwen3-4B，草稿为三层 DFlash，块大小固定为 $B=16$。$K$ 为推理批大小（请求数）；TPS 为吞吐量（token/s），$\tau$ 为平均接受长度。评测使用单张 L40S。}
\else
\caption{\textbf{\dpalRenyi{} has the highest $\tau$ at every batch size and the highest throughput in 9 of 10 settings.} 128 MATH-500 prompts on Qwen3-4B with the 100K checkpoints; $K$ is the inference batch size (number of requests) and TPS is output tokens per second.}
\fi
\label{tab:grouped}
\begin{tabular*}{\linewidth}{@{}l@{\extracolsep{\fill}\hspace{8pt}}*{4}{w{r}{33pt}@{\extracolsep{0pt}\hspace{3pt}}w{r}{17.5pt}@{\extracolsep{\fill}\hspace{6pt}}}w{r}{33pt}@{\extracolsep{0pt}\hspace{3pt}}w{r}{17.5pt}@{}}
\toprule
& \multicolumn{2}{c}{$K=1$} & \multicolumn{2}{c}{$K=2$} & \multicolumn{2}{c}{$K=4$} & \multicolumn{2}{c}{$K=8$} & \multicolumn{2}{c}{$K=16$}\\
\cmidrule(lr){2-3}\cmidrule(lr){4-5}\cmidrule(lr){6-7}\cmidrule(lr){8-9}\cmidrule(l){10-11}
Method & TPS & $\tau$ & TPS & $\tau$ & TPS & $\tau$ & TPS & $\tau$ & TPS & $\tau$\\
\midrule
\multicolumn{11}{c}{\textit{Temperature $T=0$}}\\
\midrule
\staticCE{} & 310.24 & 5.46 & 423.17 & 5.48 & 607.93 & 5.46 & 990.60 & 5.49 & 1628.15 & 5.49\\ 
\dpace{} & 320.13 & 6.06 & 466.27 & 6.06 & 669.64 & 6.04 & \textbf{1120.19} & 6.06 & 1714.12 & 6.06\\ 
\addlinespace[1.5pt]
\dpalRenyi{} & \textbf{339.73} & \textbf{6.10} & \textbf{477.46} & \textbf{6.13} & \textbf{686.75} & \textbf{6.10} & 1094.14 & \textbf{6.13} & \textbf{1797.82} & \textbf{6.11}\\ 
\midrule
\multicolumn{11}{c}{\textit{Temperature $T=1$}}\\
\midrule
\staticCE{} & 252.70 & 4.82 & 342.24 & 4.86 & 487.59 & 4.83 & 788.09 & 4.82 & 1155.27 & 4.83\\ 
\dpace{} & 274.51 & 5.21 & 362.02 & 5.25 & 515.48 & 5.23 & 786.06 & 5.29 & 1215.91 & 5.27\\ 
\addlinespace[1.5pt]
\dpalRenyi{} & \textbf{276.68} & \textbf{5.30} & \textbf{370.04} & \textbf{5.32} & \textbf{530.42} & \textbf{5.33} & \textbf{826.42} & \textbf{5.35} & \textbf{1263.31} & \textbf{5.30}\\ 
\bottomrule
\end{tabular*}
\end{table}

%% file: signal_results.tex
\centering
\normalsize
\setlength{\tabcolsep}{2pt}
\renewcommand{\arraystretch}{1.12}
\ifdefined\ChineseDraft
\caption{\raggedright 两种位置权重下的局部损失对比。每组固定权重信号和平滑系数，仅将 CE 换成 Rényi；数字为平均 $\tau$。}
\else
\caption{\raggedright\textbf{R\'enyi beats CE under both confidence and overlap weights.} Each pair fixes the weight signal and floor and changes only the loss. Confidence with CE is \dpace{}; overlap gives \dpalCE{} and \dpalRenyi{}. Mean $\tau$ over six benchmarks.}
\fi
\label{tab:signal}
\begin{tabular*}{\linewidth}{@{\extracolsep{\fill}}llrrr@{}}
\toprule
Loss & Signal & $\rho$ & $T{=}0$ & $T{=}1$\\
\midrule
CE & Confidence & 0.5 & 3.50 & 3.36\\ 
R\'enyi & Confidence & 0.5 & \textbf{3.58} & \textbf{3.42}\\ 
\midrule
CE & Overlap & 0.3 & 3.50 & 3.36\\ 
R\'enyi & Overlap & 0.3 & \textbf{3.59} & \textbf{3.43}\\ 
\bottomrule
\end{tabular*}

%% file: dynamic_control_results.tex
\centering\normalsize
\renewcommand{\arraystretch}{1.12}
\setlength{\tabcolsep}{3pt}
\ifdefined\ChineseDraft
\caption{\raggedright Qwen3-4B、三层 DFlash、$B=16$ 下的位置权重对照（平均 $\tau$）。静态指数权重取 $\gamma=7$；\dpalname{} 使用式~\eqref{eq:weights}，取 $\rho=0.5$。}
\else
\caption{\raggedright\textbf{\dpalname{} weights beat fixed decay for all losses from baselines.} KL+TV follows the adaptive mixture of LK Losses with per-position $\lambda_t$. Static uses DFlash's decay $e^{-(t-1)/7}$; \dpalname{} uses $\rho=0.3$ ($0.5$ for CE). Mean $\tau$ over six benchmarks.}
\fi
\label{tab:dynamic-controls}
\begin{tabular*}{\linewidth}{@{\extracolsep{\fill}}lrrrr@{}}
\toprule
& \multicolumn{2}{c}{$T=0$} & \multicolumn{2}{c}{$T=1$}\\
\cmidrule(lr){2-3}\cmidrule(l){4-5}
Local loss & Static & \dpalname{} & Static & \dpalname{}\\
\midrule
CE & 3.24 & \textbf{3.50} & 3.11 & \textbf{3.36}\\ 
LA & 3.27 & \textbf{3.56} & 3.15 & \textbf{3.40}\\ 
CE+TV & 3.40 & \textbf{3.48} & 3.27 & \textbf{3.34}\\ 
KL+TV & 3.20 & \textbf{3.53} & 3.08 & \textbf{3.37}\\ 
\bottomrule
\end{tabular*}

%% file: capacity_results.tex
\begin{table}[t]
\begin{minipage}[t]{0.485\linewidth}
\vspace{-20pt}
\centering
\small
\setlength{\tabcolsep}{2pt}
\renewcommand{\arraystretch}{1.12}
\ifdefined\ChineseDraft
\caption{\raggedright Qwen3-4B 的块大小消融（平均 $\tau$）。草稿固定为三层。}
\else
\caption{\raggedright\textbf{\dpalRenyi{} has the highest $\tau$ at every block size.} Mean $\tau$ over six benchmarks; three-layer drafts.}
\fi
\label{tab:block-size}
\begin{tabular*}{\linewidth}{@{\extracolsep{\fill}}r*{4}{r}@{}}
\toprule
& \multicolumn{2}{c}{\dpace{}} & \multicolumn{2}{c}{\dpalRenyi{}}\\
\cmidrule(lr){2-3}\cmidrule(l){4-5}
$B$ & $T{=}0$ & $T{=}1$ & $T{=}0$ & $T{=}1$\\
\midrule
8 & 3.41 & 3.28 & \textbf{3.48} & \textbf{3.34}\\ 
12 & 3.51 & 3.36 & \textbf{3.57} & \textbf{3.42}\\ 
16 & 3.50 & 3.36 & \textbf{3.58} & \textbf{3.41}\\ 
\bottomrule
\end{tabular*}
\end{minipage}\hfill
\begin{minipage}[t]{0.485\linewidth}
\vspace{0pt}
\centering
\small
\setlength{\tabcolsep}{2pt}
\renewcommand{\arraystretch}{1.12}
\ifdefined\ChineseDraft
\caption{\raggedright Qwen3-4B 的草稿深度消融（平均 $\tau$）。固定 $B=16$，两种深度使用相同的三层目标特征。}
\else
\caption{\raggedright\textbf{Deeper drafts help both methods; \dpalRenyi{} leads at both depths.} Mean $\tau$ over six benchmarks.}
\fi
\label{tab:draft-depth}
\begin{tabular*}{\linewidth}{@{\extracolsep{\fill}}r*{4}{r}@{}}
\toprule
& \multicolumn{2}{c}{\dpace{}} & \multicolumn{2}{c}{\dpalRenyi{}}\\
\cmidrule(lr){2-3}\cmidrule(l){4-5}
Layers & $T{=}0$ & $T{=}1$ & $T{=}0$ & $T{=}1$\\
\midrule
3 & 3.50 & 3.36 & \textbf{3.58} & \textbf{3.41}\\ 
5 & 3.75 & 3.56 & \textbf{3.81} & \textbf{3.63}\\ 
\bottomrule
\end{tabular*}
\end{minipage}
\end{table}

%% file: smoothing_results.tex
\centering
\normalsize
\setlength{\tabcolsep}{2pt}
\renewcommand{\arraystretch}{1.12}
\ifdefined\ChineseDraft
\caption{\raggedright \dpalRenyi{} 平滑下限消融（平均 $\tau$），默认 $\rho=0.5$。各列固定损失、目标与块大小。Llama 为 Instruct 版本。}
\else
\caption{\raggedright\textbf{Without the floor, \dpalRenyi{} collapses; $\rho\in\{0.3,0.5\}$ works best.} Mean $\tau$ over six benchmarks as the floor $\rho$ varies.}
\fi
\label{tab:smoothing}
\begin{tabular*}{\linewidth}{@{\extracolsep{\fill}}l*{6}{r}@{}}
\toprule
& \multicolumn{4}{c}{Qwen3-4B} & \multicolumn{2}{c}{Llama-3.1-8B}\\
\cmidrule(lr){2-5}\cmidrule(l){6-7}
& \multicolumn{2}{c}{$B=8$} & \multicolumn{2}{c}{$B=16$} & \multicolumn{2}{c}{$B=10$}\\
\cmidrule(lr){2-3}\cmidrule(lr){4-5}\cmidrule(l){6-7}
$\rho$ & $T{=}0$ & $T{=}1$ & $T{=}0$ & $T{=}1$ & $T{=}0$ & $T{=}1$\\
\midrule
0 & 1.51 & 1.50 & 1.03 & 1.03 & 1.60 & 1.51\\ 
0.3 & \textbf{3.51} & \textbf{3.37} & \textbf{3.59} & \textbf{3.43} & 3.29 & 2.69\\ 
0.5 & 3.48 & 3.34 & 3.58 & 3.41 & \textbf{3.31} & \textbf{2.73}\\ 
0.7 & 3.43 & 3.30 & 3.50 & 3.35 & 3.23 & 2.65\\ 
0.9 & 3.37 & 3.25 & 3.36 & 3.24 & 3.09 & 2.58\\ 
\bottomrule
\end{tabular*}

%% file: target_results.tex
\centering
\small
\setlength{\tabcolsep}{2pt}
\renewcommand{\arraystretch}{1.12}
\ifdefined\ChineseDraft
\caption{\raggedright 8B 目标模型对照（平均 SR 与 $\tau$）。草稿均为三层，同一目标下共享语料和评测设置。Llama 为 Instruct 版本。}
\else
\caption{\raggedright\textbf{\dpalRenyi{} leads on both 8B targets in SR and $\tau$.} Means over six benchmarks; three-layer drafts.}
\fi
\label{tab:targets}
\begin{tabular*}{\linewidth}{@{\extracolsep{\fill}}ll*{4}{r}@{}}
\toprule
& & \multicolumn{2}{c}{\shortstack{Llama-3.1-8B\\$B=10$}} & \multicolumn{2}{c}{\shortstack{Qwen3-8B\\$B=16$}}\\
\cmidrule(lr){3-4}\cmidrule(l){5-6}
Method & Metric & $T{=}0$ & $T{=}1$ & $T{=}0$ & $T{=}1$\\
\midrule
\dpace{} & SR & 2.47 & 1.77 & 2.59 & 2.37\\
 & $\tau$ & 3.18 & 2.61 & 3.54 & 3.34\\ 
\addlinespace[1.5pt]
\dpalRenyi{} & SR & \textbf{2.57} & \textbf{1.84} & \textbf{2.65} & \textbf{2.40}\\
 & $\tau$ & \textbf{3.31} & \textbf{2.73} & \textbf{3.63} & \textbf{3.39}\\ 
\bottomrule
\end{tabular*}

%% file: thinking_results.tex
\begin{table}[t]
\vspace{-28pt}
\centering\normalsize
\setlength{\tabcolsep}{2pt}
\renewcommand{\arraystretch}{1.0}
\caption{\textbf{With thinking enabled, \dpalRenyi{} leads mean $\tau$ at $T=0$ and \dpalKL{} at $T=1$.} DFlash is the reference checkpoint (protocol in Appendix~\ref{app:thinking}).}
\label{tab:thinking}
\small
\begin{tabular*}{\linewidth}{@{\extracolsep{\fill}}lrrrrrrr@{}}
\toprule
& \multicolumn{7}{c}{Acceptance length $\tau$}\\
\cmidrule(l){2-8}
Method & GSM8K & MATH-500 & HumanEval & MBPP & MT-Bench & Alpaca & Mean\\
\midrule
\multicolumn{8}{c}{\textit{Temperature $T=0$}}\\
\midrule
DFlash & 3.64 & 3.53 & 3.92 & 3.55 & 3.12 & 3.04 & 3.47\\ 
\dpace{} & 3.94 & 3.82 & 4.14 & 3.96 & 3.28 & 3.17 & 3.72\\ 
\dpalLA{} & 3.96 & \textbf{3.91} & \textbf{4.21} & 3.85 & 3.47 & 3.22 & 3.77\\ 
\dpalKL{} & 3.98 & 3.90 & 4.20 & 3.96 & 3.46 & 3.23 & 3.79\\ 
\dpalRenyi{} & \textbf{4.09} & 3.87 & 4.05 & \textbf{4.02} & \textbf{3.52} & \textbf{3.31} & \textbf{3.81}\\ 
\midrule
\multicolumn{8}{c}{\textit{Temperature $T=1$}}\\
\midrule
DFlash & 3.21 & 3.10 & 3.08 & 3.01 & 2.69 & 2.62 & 2.95\\ 
\dpace{} & 3.42 & 3.34 & 3.29 & 3.20 & 2.84 & 2.73 & 3.14\\ 
\dpalLA{} & 3.44 & 3.35 & 3.29 & 3.18 & 2.86 & 2.73 & 3.14\\ 
\dpalKL{} & \textbf{3.47} & \textbf{3.38} & \textbf{3.33} & \textbf{3.23} & \textbf{2.87} & \textbf{2.79} & \textbf{3.18}\\ 
\dpalRenyi{} & \textbf{3.47} & 3.37 & 3.31 & 3.21 & \textbf{2.87} & 2.77 & 3.17\\ 
\bottomrule
\end{tabular*}
\end{table}

%% file: experiment_details.tex
\ifdefined\ChineseDraft
\section{实验设置与损失定义}\label{app:experiments}
\subsection{训练与评测}
表~\ref{tab:results} 使用 Qwen3-4B 语料中的 99,171 条有效训练序列，表~\ref{tab:grouped} 沿用这些 checkpoint；其他受控实验由 SpecForge 随机选取约36K条 ShareGPT 提示。每个冻结目标生成回复的温度为0.7，Qwen3 使用非思考模式。同一比较中的方法共享数据。表~\ref{tab:settings} 汇总其余设置；表~\ref{tab:results}--\ref{tab:grouped} 的结果均为三个评测随机种子的平均值。
\else
\section{Experimental Settings and Loss Definitions}\label{app:experiments}
\subsection{Training and evaluation}
Table~\ref{tab:results} uses the Qwen3-4B-Instruct-100K dataset released by the DFlash authors, which \dpace{} also uses for its main Qwen3-4B experiments~\citep{chen2026dflash,wu2026dpace}. Of its 99,987 rows, 99,171 are valid training sequences. We randomly sample ShareGPT prompts with SpecForge for the roughly 36K-example runs. All training responses in this paper are generated by the frozen target at temperature 0.7; Qwen3 uses non-thinking mode. Methods share data within each comparison (settings in Table~\ref{tab:settings}). Results in Tables~\ref{tab:results}--\ref{tab:grouped} average evaluation seeds 980406, 980407, and 980408. Across evaluation seeds, the standard deviation of each six-benchmark mean $\tau$ is 0 at $T=0$ and at most 0.02 at $T=1$. Every configuration uses training seed 42.
\fi

\begin{table}[!htbp]
\centering\normalsize
\renewcommand{\arraystretch}{1.13}
\ifdefined\ChineseDraft
\caption{训练与评测设置。块大小包含一个不参与损失的 anchor；不同草稿深度使用相同的三个目标特征层。}
\else
\caption{\textbf{Training and evaluation settings.} Each block includes one anchor token excluded from the loss. Both draft depths use the same three target feature layers.}
\fi
\label{tab:settings}
\begin{tabular}{@{}p{0.29\linewidth}p{0.67\linewidth}@{}}
\toprule
\multicolumn{2}{l}{\textit{\ifdefined\ChineseDraft 训练\else Training\fi}}\\
\ifdefined\ChineseDraft 优化器与轮数\else Optimizer / epochs\fi & AdamW / 6\\
\ifdefined\ChineseDraft 学习率与调度\else Learning rate / schedule\fi & $6\times10^{-4}$; \ifdefined\ChineseDraft 余弦衰减，4\% warm-up\else cosine decay, 4\% warm-up\fi\\
\ifdefined\ChineseDraft 全局 batch / seed\else Global batch / seed\fi & \ifdefined\ChineseDraft 表~\ref{tab:results} 与表~\ref{tab:grouped} 为8，其余为4；seed 42\else 8 (Tables~\ref{tab:results} and~\ref{tab:grouped}), 4 (other runs); seed 42\fi\\
\ifdefined\ChineseDraft 序列与 anchor 上限\else Sequence / anchor limits\fi & 3072 tokens / 512 anchors\\
\ifdefined\ChineseDraft 默认块大小\else Default block size\fi & Qwen: $B=16$; Llama: $B=10$\\
\ifdefined\ChineseDraft 损失分布温度\else Loss temperature\fi & 1 \ifdefined\ChineseDraft （目标与草稿）\else (target and draft)\fi\\
\midrule
\multicolumn{2}{l}{\textit{\ifdefined\ChineseDraft 评测\else Evaluation\fi}}\\
\ifdefined\ChineseDraft 后端\else Backend\fi & SGLang 0.5.19 DFLASH\\
\ifdefined\ChineseDraft 评测温度\else Evaluation temperatures\fi & $T=0$, $T=1$\\
\ifdefined\ChineseDraft 生成长度上限\else Generation limit\fi & 2048 \ifdefined\ChineseDraft 个新 token\else new tokens\fi\\
\bottomrule
\end{tabular}
\end{table}

\ifdefined\ChineseDraft
\paragraph{目标与草稿分布。}
我们将缓存的目标最终 hidden states 输入冻结的目标输出头，获得完整词表 logits，并通过 softmax 得到 $p_t$。草稿在 anchor 上下文与 masked block 输入下预测相应位置，通过 softmax 得到 $q_t$。两者按记录序列中的预测位置对齐；并行草稿不接收目标在后续位置可见的完整前缀。损失中的两种 softmax 温度均为1，记录标签则来自温度0.7的生成分布。

\paragraph{推理协议。}
SGLang 0.5.19 DFLASH 在两种评测温度下均通过 argmax 生成草稿候选。$T=0$ 时，验证器保留与目标 argmax 一致的前缀；$T=1$ 时，后端对这些确定性提案执行 target-only 随机接受与拒绝校正。评测不使用 top-$k$ 或 top-$p$ 截断，两个接受阈值均为1。实际提案分布是草稿 argmax 上的点质量，而不是训练 softmax $q_t$。因此，训练重叠在两种设置下都是代理信号；报告的 $\tau$ 来自实际验证过程，而非用重叠估算。

\paragraph{评测子集。}
GSM8K、MATH-500 和 MBPP 各使用128条，HumanEval 使用164条，MT-Bench 使用80条首轮提示；子集 seed 为42。Alpaca 使用固定的128条单轮提示，列表随补充材料提供于 \texttt{alpaca\_128.jsonl}。同一目标下，各方法与温度共享提示。

\subsection{局部损失与位置权重}\label{app:losses}
所有目标采用式~\eqref{eq:objective} 的加权求和与平均方式，只替换局部损失或权重信号。表~\ref{tab:loss-definitions} 给出所测试的局部损失，其中 $y_t^\star$ 为记录序列中的目标 token，$\alpha_t=1-\TV(p_t,q_t)$。散度与重叠均在完整词表上计算。
\else
\paragraph{Compute.}
We estimate that running all training experiments on NVIDIA H200 GPUs would require approximately 840 GPU-hours, including 138 GPU-hours for the eight 100K runs. We run all evaluations on NVIDIA L40S GPUs.

\paragraph{Target and draft distributions.}
The frozen target output head maps cached final hidden states to full-vocabulary logits; softmax gives $p_t$. At each recorded position, the draft predicts $q_t$ from the anchor context and a masked block. It does not see the full prefix available to the target at later positions. Both loss softmaxes use temperature one; we generate labels at temperature 0.7.

\paragraph{Inference protocol.}
SGLang 0.5.19 DFLASH uses argmax draft proposals at both temperatures. At $T=0$, verification retains the target-argmax prefix; at $T=1$, it applies target-only stochastic acceptance and rejection correction, which accepts the deterministic proposal $\hat y_t$ with probability $p_t(\hat y_t)$~\citep{sglang2026dflash}. We disable top-$k$ and top-$p$ truncation and set SGLang's single-token and accumulated acceptance thresholds to one. We measure $\tau$ through verification; overlap serves as a training proxy.

The grouped-request experiment in Table~\ref{tab:grouped} uses 128 fixed MATH-500 prompts and one L40S. All inference batch sizes, including $K=1$, use grouped-request inference, so its throughput is not directly comparable with the single-request results in Table~\ref{tab:results}. It evaluates the 100K checkpoints of \staticCE{}, \dpace{}, and \dpalRenyi{} from Table~\ref{tab:results}.

\paragraph{Evaluation subsets.}
We use 128 examples each from GSM8K, MATH-500, and MBPP, 164 from HumanEval, and 80 first-turn MT-Bench prompts, with subset seed 42. Alpaca uses a fixed list of 128 single-turn prompts, supplied as \texttt{alpaca\_128.jsonl}. We use the same prompts across methods and temperatures within each target.

\Needspace{6\baselineskip}
\paragraph{Thinking-mode comparison.}\label{app:thinking}
Table~\ref{tab:thinking} uses a separate set of three-layer, $B=16$ checkpoints on the same six benchmark subsets, with thinking enabled. The \dpace{} and \dpalname{} checkpoints use $\rho=0.3$. The context limit is 16384 tokens and the generation limit is 8192; both reasoning and final-answer tokens count toward the metrics. Each run uses one L40S and one active request. We retain responses that reach the generation limit.

\subsection{Local losses and position weights}\label{app:losses}
Table~\ref{tab:loss-definitions} lists the local fitting terms, with recorded target token $y_t^\star$ and training overlap $\widehat\alpha_t=1-\TV(p_t,q_t)$. All \dpalname{} objectives use the detached weights in Equation~\eqref{eq:weights} and differ only in their local loss. Divergences and overlap use the full vocabulary.
\fi

\ifdefined\ChineseDraft
\begin{table}[!htbp]
\else
\begin{table}[H]
\fi
\centering\normalsize
\renewcommand{\arraystretch}{1.2}
\ifdefined\ChineseDraft
\caption{实验使用的局部损失。本文的 LK hybrid 变体在每个块、每个位置使用 $\lambda_t=\exp(-3\alpha_t)$，求导时将该系数视为常数。}
\else
\caption{\textbf{Local losses compared in this paper.} \dpalKLTV{} computes $\lambda_t=\exp(-3\widehat\alpha_t)$ for each block and position and detaches it during optimization.}
\fi
\label{tab:loss-definitions}
\begin{tabular}{@{}ll@{}}
\toprule
\ifdefined\ChineseDraft 目标\else Objective\fi & \ifdefined\ChineseDraft 局部损失 $\ell_t$\else Local loss $\ell_t$\fi\\
\midrule
Cross-entropy & $\ell_{\mathrm{CE},t}=-\log q_t(y_t^\star)$\\
\ifdefined\ChineseDraft\else \dpalLA{} & $-\log[1-\TV(p_t,q_t)]$\\\fi
\ifdefined\ChineseDraft\else \dpalKL{} & $\KL(p_t\Vert q_t)$\\\fi
\dpalRenyi{} & $D_{1/2}(p_t\Vert q_t)$\\
\dpalTV{} & $\TV(p_t,q_t)$\\
\dpalCETV{} & $0.1\ell_{\mathrm{CE},t}+0.9\TV(p_t,q_t)$\\
\dpalKLTV{} & $\lambda_t\KL(p_t\Vert q_t)+(1-\lambda_t)\TV(p_t,q_t)$\\
\bottomrule
\end{tabular}
\end{table}

\ifdefined\ChineseDraft
\paragraph{TV 加权与长度目标。}
D-PATV 将局部 TV 损失乘以式~\eqref{eq:weights} 中经过平滑并 detach 的权重 $w_t=\widetilde R_t\widetilde F_t$。第~\ref{sec:position-training} 节长度训练损失 $\mathcal L_{\mathrm{len}}$ 的精确 TV 梯度系数则为 $R_{t-1}F_t/D$（附录~\ref{app:position}）。因此，这一对照并未复现 Bebop 的损失。

\paragraph{混合损失。}
DSpark 使用 $0.1\mathcal L_{\mathrm{CE}}+0.9\|p-q\|_1$，其中 $\|p-q\|_1=2\TV$；本文 CE+TV 则使用表中的 $0.1\ell_{\mathrm{CE},t}+0.9\TV(p_t,q_t)$。LK hybrid 的局部形式为
\begin{equation}
 \mathcal L_{\mathrm{LK}}^\lambda
 =\lambda\KL(p\Vert q)+(1-\lambda)\TV(p,q).
 \label{eq:mixtures}
\end{equation}
本文分别计算每个块、每个位置的 $\lambda_t=\exp(-3\alpha_t)$，求导时固定该系数。原 LK Losses 先对序列与 batch 聚合接受率，再计算位置系数~\citep{samarin2026lk}；本文不做这一聚合。

\paragraph{位置权重。}
静态权重为 $w_t=\exp(-(t-1)/\gamma)$，其中 $t=1$ 是第一个预测位置；$B=8,10,12,16$ 分别使用 $\gamma=4,5,6,7$。动态权重采用式~\eqref{eq:weights}：D-PACE 用 $q_t(y_t^\star)$ 计算权重，D-PA 变体使用 $\alpha_t$。表~\ref{tab:results} 中 D-PACE 取 $\rho=0.5$，D-PA 取 $\rho=0.3$；其他消融默认取 $\rho=0.5$，表~\ref{tab:signal} 单独列出比较值。所有损失先在序列内对有效块取平均，再对序列取平均，不按位置权重之和归一化。
\else
DSpark's $0.1\mathcal L_{\mathrm{CE}}+0.9\|p-q\|_1$ has twice the TV coefficient of our CE+TV loss. \dpalKLTV{} computes $\lambda_t$ at each position, whereas LK Losses aggregates acceptance across sequence and batch before setting its hybrid coefficient~\citep{samarin2026lk}.

Static weights use $\exp(-(t-1)/\gamma)$, with $\gamma=4,5,6,7$ for $B=8,10,12,16$. \dpace{} smooths confidence with $\rho=0.5$. \dpalname{} smooths $\widehat\alpha_t$ with $\rho=0.3$ in Table~\ref{tab:results} and with $\rho=0.5$ in the ablations unless varied; \dpalCE{} uses $\rho=0.3$. We average blocks within sequences, then sequences, without normalizing by the weight sum.

\paragraph{Implementation.}
The two weight scans cost $O(D_b)$; vocabulary sums cost $O(D_bV)$. We compute the R\'enyi loss in 32-bit floating point (FP32) using log-softmax and logsumexp.

\paragraph{Gradient conflict in TV mixtures.}
A fixed positive TV coefficient leaves a mixture nondifferentiable, and the auxiliary gradient can oppose the TV gradient when the terms share parameters~\citep{du2018adapting,yu2020gradient}. Let $g_T=\nabla_\theta\TV$ and $g_F=\nabla_\theta\mathcal L_F$. At differentiable points, a stochastic gradient descent step of size $\eta$ on $a\TV+b\mathcal L_F$ changes TV by
\begin{equation}
 \delta\TV=-\eta\bigl(a\|g_T\|_2^2+b\,g_T^\top g_F\bigr)+o(\eta),
 \label{eq:conflict}
\end{equation}
so the mixed step raises TV to first order when $a\|g_T\|_2^2+b\,g_T^\top g_F<0$.

For target $p=(0.45,0.35,0.20)$ and draft $q=(0.55,0.01,0.44)$, vary two draft logits and fix the third at zero. With hard-target CE for token 1 and $(a,b)=(0.9,0.1)$, $g_T^\top(0.1g_{\mathrm{CE}}+0.9g_T)<0$: descent on the mixed loss raises TV to first order (Figure~\ref{fig:conflict-toy}).
\input{conflict_figure}

KL and TV gradients can also oppose, which can affect \dpalKLTV{}. For $p=(0.4,0.5,0.1)$ and $q(\theta)=\operatorname{softmax}(\log(0.1,0.6,0.3)+\theta(0,1,-1))$, $d\TV/d\theta=0.03$ and $d\KL/d\theta=-0.10$ at $\theta=0$: the KL and TV gradients oppose. With detached position weights, Equation~\eqref{eq:conflict} applies to the weighted sum of these gradients over positions.

\fi
\FloatBarrier

%% file: conflict_figure.tex
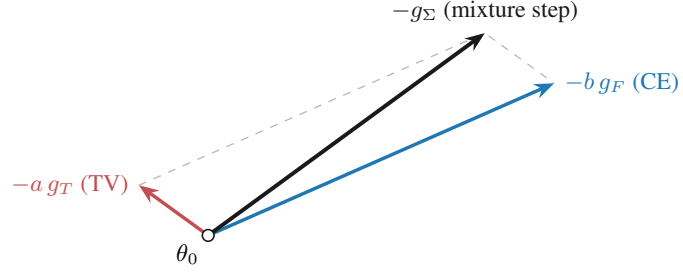
\begin{figure}[!htbp]
\centering
\definecolor{ceBlue}{RGB}{31,119,180}
\definecolor{tvRed}{RGB}{196,78,82}
\begin{tikzpicture}[x=1.45cm,y=1.45cm,>={Stealth[length=2.6mm,width=2mm]}]
\coordinate (O) at (0,0);
\coordinate (F) at (3.157,1.388);
\coordinate (T) at (-0.632,0.458);
\coordinate (S) at (2.524,1.846);
\draw[dashed,gray!55,line width=.5pt] (T) -- (S) -- (F);
\draw[->,line width=1.3pt,ceBlue] (O) -- (F) node[right,font=\small,text=ceBlue]{$-b\,g_F$ (CE)};
\draw[->,line width=1.3pt,tvRed] (O) -- (T) node[left,font=\small,text=tvRed]{$-a\,g_T$ (TV)};
\draw[->,line width=1.5pt,gray!20!black] (O) -- (S) node[above,font=\small]{$-g_\Sigma$ (mixture step)};
\fill[white,draw=gray!20!black,line width=.6pt] (O) circle (2.2pt);
\node[below left,font=\small] at (O) {$\theta_0$};
\end{tikzpicture}
\caption{\textbf{A CE+TV descent step can raise TV.} Negative weighted gradients of TV ($a=0.9$) and CE ($b=0.1$) at one parameter point of the three-token example, on one shared scale and rotated $25^\circ$ for display. The two gradients have a negative inner product, and the mixture step $-g_\Sigma$ raises TV to first order ($g_T^\top g_\Sigma<0$).}
\label{fig:conflict-toy}
\end{figure}